%% file: main.tex
\documentclass[a4paper]{article}
\usepackage{INTERSPEECH_v2}
\usepackage{amsfonts, amsmath, graphicx, multirow, amsthm, amssymb, bbm, color}
\input{definitions}
\graphicspath{{./img/}}

\title{Learning Similarity Functions for Pronunciation Variations}
\name{Einat Naaman, Yossi Adi, and Joseph Keshet}
\address{Department of Computer Science, Bar-Ilan University, Ramat-Gan, Israel}
\email{jkeshet@cs.biu.ac.il}

\begin{document}
\maketitle
\begin{abstract}
A significant source of errors in Automatic Speech Recognition (ASR) systems is due to pronunciation variations which occur in spontaneous and conversational speech. Usually ASR systems use a finite lexicon that provides one or more pronunciations for each word. In this paper, we focus on learning a similarity function between two pronunciations. The pronunciations can be the canonical and the surface pronunciations of the same word or they can be two surface pronunciations of different words. This task generalizes problems such as lexical access (the problem of learning the mapping between words and their possible pronunciations), and defining word neighborhoods. It can also be used to dynamically increase the size of the pronunciation lexicon, or in predicting ASR errors. We propose two methods, which are based on recurrent neural networks, to learn the similarity function. The first is based on binary classification, and the second is based on learning the ranking of the pronunciations. We demonstrate the efficiency of our approach on the task of lexical access using a subset of the Switchboard conversational speech corpus. Results suggest that on this task our methods are superior to previous methods which are based on graphical Bayesian methods.
\end{abstract}

\label{intro}
\input{01_intro}

\label{prob_set}
\input{02_prob-set}

\label{model}
\input{03_model}

\label{exp}
\input{04_exp}

\label{analysis}
\input{05_ana}

\label{conclusion}
\input{06_conc}

\bibliographystyle{IEEEtran}
\bibliography{ref}

\end{document}

%% file: definitions.tex
\usepackage{bm}
\usepackage{amssymb,amsmath,graphicx,bbm,color,booktabs}
\usepackage{amsopn}

\newcommand{\vp}{\bm{p}}

\newcommand{\vu}{\bm{u}}               
\newcommand{\vv}{\bm{v}}               
\newcommand{\vw}{\bm{w}}

    \newcommand{\Pc}{\mathcal{P}}

\newcommand{\R}{\mathbb{R}}

\DeclareMathOperator*{\argsort}{arg\,sort^k}

\renewcommand{\eqref}[1]{Eq.~(\ref{#1})}

\renewcommand{\P}{\mathcal{P}^*}

%% file: 01_intro.tex
\section{Introduction}

Spontaneous and conversational speech are significantly different both acoustically and linguistically from read speech. One of the key differences is the vast pronunciation variations in spontaneous and conversational speech, as the speaking rate is accelerated and consequently the pronunciation becomes reduced or coarticulated. Other factors, such as the neighboring words and the speaker's style, also influence the way words are produced. 

We distinguish between two types of pronunciations of a word. The typical pronunciation found in a dictionary is called \emph{canonical} pronunciation, whereas the actual way in which speakers produce the word is called \emph{surface} pronunciation. Spontaneous speech often includes pronunciations that differ from the one found in the dictionary. For example, pronunciations of the word ``probably'' in the Switchboard conversational speech corpus include [p r aa b iy], [p r aa l iy], [p r ay], etc. \cite{greenberg:stp}. Fewer than half of the word productions are pronounced canonically in the phonetically transcribed portion of Switchboard \cite{livescu_thesis}. 

In this work we propose to learn a similarity function between two pronunciation variations. The function should score the similarity between any type of pronunciations. The input can be, for example, canonical and surface pronunciations or it can be two surface pronunciations. We expect such a function to output a high number if the inputs are canonical and surface pronunciations of the same word or if the input consists of two surface pronunciations of the same word. 

Such a similarity measure can be utilized in many tasks, including lexical access, word neighborhood and pronunciation scoring. In the task of lexical access, the goal is to predict which word in the dictionary was uttered given its pronunciation in terms of sub-word units \cite{FisEtal1989,Jyothi2011,hao2012discriminative}. The problem of lexical access can be handled using a pronunciation similarity function as follows: the input surface pronunciation is compared to all of the (canonical) pronunciations in the dictionary using the similarity function, and the one with the maximal similarity is predicted as the articulated word.

In the word neighborhood task the goal is to find the acoustic neighborhood density of a given word \cite{JyothiLivescu2014}. It has been suggested as an explanatory variable for quantifying errors in ASR, but is also used in linguistic studies. In most works in the psycholinguistics literature, word neighborhood is defined to be the set of words which differ by a single phone from the given word. Here we propose a different definition which is closer to the definition suggested in \cite{JyothiLivescu2014}. The word neighborhood can be defined as all the  words in the dictionary in proximity to the given word, where proximity is measured using the similarity function between pronunciations.

ASR systems are based on a finite dictionary which provides each word with one or more pronunciations. The variance of pronunciations in spontaneous and conversational speech leads to a high rate of errors in ASR systems \cite{Jurafsky-triphones, SaraclarKhudanpur2004}. The standard approach to this problem is to expand the dictionary, either by adding alternate pronunciations with probabilities, or with phonetic substitution, insertion and deletion rules derived from linguistic knowledge and learned from data. A similarity function can be used to add pronunciation variations to each word either statically or dynamically during the dictionary lookup.

Mapping between words and their possible pronunciations in terms of sub-word units was explored in light of the lexical access task \cite{livescu2004feature,Jyothi2011,hao2012discriminative}. Note that this problem is different from the grapheme-to-phoneme problem, in which pronunciations are predicted from a word's spelling, whereas in lexical access we assume a dictionary of canonical pronunciations like the one used in speech recognition. Learning word neighborhood automatically was proposed by \cite{JyothiLivescu2014}. 

Our work is restricted to proposing algorithms for learning the pronunciation similarity score. Specifically we propose two different deep network architectures to tackle this problem. The first is based on binary classification of two recurrent neural networks (RNNs), while the second is based on triplet networks designed to rank the pronunciations.

The paper is organized as follows. Section~\ref{sec:prob-set} introduces our notation and the problem definition. In Section~\ref{sec:model} we propose two deep network architectures to learn similarity between two pronunciations. In Section~\ref{sec:exp} we present a set of experiments on the Switchboard conversational speech corpus, and in Section~\ref{sec:ana} we analyze the experimental results. We conclude the paper in Section~\ref{sec:conc}.

%% file: 02_prob-set.tex

\section{Problem settings}\label{sec:prob-set}

We denote a word pronunciation by a sequence of phones, $\vp = (p_1,\ldots,p_N)$ where $p_n \in \Pc$ for all $1 \le n \le N$ and $\Pc$ is the set of all sub-word units (all phones). Naturally, $N$ is not fixed since the number of phonemes varies across different words. We denote the set of all finite-length phone streams as $\P$.

Given two pronunciation sequences, our goal is to find a similarity function between these two sequences. Formally, our goal is to learn a function $f: \P \times \P \rightarrow  \mathbb{R}$ which gets as input two pronunciation sequences and returns the similarity between the two. That is if two pronunciations $\vp_1, \vp_2 \in \P$ are similar, then the function $f(\vp_1, \vp_2)$ will be high. Otherwise it will be low.

This type of function can be utilized in various tasks. For example, in the lexical access task \cite{hao2012discriminative} the goal is to predict which word $w$ from a  finite dictionary $\mathcal{V}$ is associated with a given surface pronunciation $\vp^{s}$. Assume that the dictionary $\mathcal{V}$ is a list of pairs, where each pair is composed of a word $w$ and its canonical pronunciation $\vp^{c}$, namely $(w,\vp^{c})\in\mathcal{V}$. Define $sort^k$ as a function that gets a set of unordered scores and returns a vector of the top $k$ maximal ordered scores. Similarly define $arg\, sort^k$ to be the function the returns the indices of the ordered scores. Then the best $k$-words can be found to be those whose canonical pronunciations get the highest similarity to the input surface pronunciation:
\begin{equation}\label{eq:comp_pronunciations}
	\vw = \argsort_{(w, \vp^c)\in\mathcal{V}} f(\vp^{s}, \vp^{c}),
\end{equation}
where $\vw$ is a list of $k$-best words in the dictionary. Either the word with the highest unigram probability can be predicted or it can be combined to generated a prediction based on the n-gram probability.

In the word neighborhood task the goal is to find the acoustic neighborhood density of a given word \cite{JyothiLivescu2014}. This can be achieved by comparing the similarity of a canonical pronunciation of a given word to the set of all canonical pronunciations in the dictionary. Formally we propose to define word neighborhood $\mathcal{N}(w)$ for a given word $w$ as the set of all words $u$ in the dictionary $\mathcal{V}$ for which the similarity function is less than some threshold $\theta$:
\begin{equation}
\mathcal{N}(w) = \{ u ~|~ \forall (u,\vp^u)\in\mathcal{V}, ~ f(\vp^w,\vp^u) < \theta \},
\end{equation}
where $\vp^w, \vp^u \in\P$ are the canonical pronunciations of the word $w$ and $u$ respectively.



%% file: 03_model.tex
\section{Network architecture}\label{sec:model}

Is this section we suggest two network architectures to learn the similarity function. Both architectures are based on Recurrent Neural Networks (RNNs) \cite{elman1990finding}. The first one is based on Siamese RNNs which were  designed as a binary classifier and were trained to minimize the negative log likelihood loss function. The second architecture is based on three identical RNNs, which are combined together to create a ranking architecture and optimized with a ranking loss function. We also tried several sequence-to-sequence architectures \cite{sutskever2014sequence, yao2015sequence}, but since all of them led to poor results in terms of Word Error Rate (WER), we will not discuss them in the paper.

\subsection{Binary loss function}

The first architecture is a network that is designed to learn a mapping between two pronunciations using a binary classifier, which is trained to predict whether two pronunciations are of the same word. Each example in the training set of $m$ examples, $S=\{(\vp_i^s, \vp_i^c, y_i)\}_{i=1}^{m}$, is composed of a surface pronunciation $\vp^s\in\P$, a canonical pronunciation $\vp^c\in\P$ and a binary label $y \in \{-1, +1\}$, which indicates if both pronunciations are of the same word or not. 

\begin{figure}[t]
  \centering
  \includegraphics[width=52mm, scale=0.8]{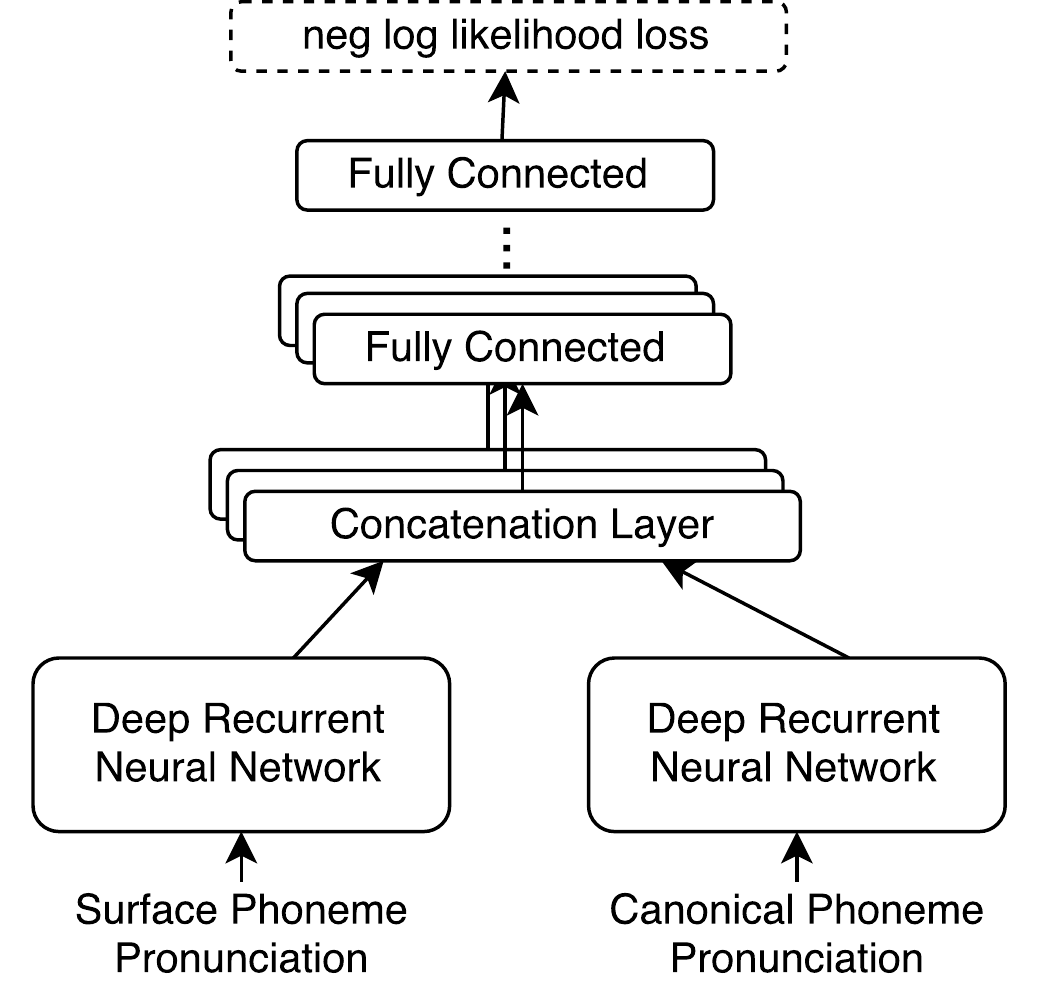}
  \caption{Network architecture of the binary loss network. The two RNNs share the same parameters.}
  \label{fig:net_arch_1}
\end{figure}

We would like to train a neural network to learn this binary mapping. In order to do so, we encode both pronunciations using two RNNs with a shared set of parameters. The input to each of the RNNs is a sequence of phones, which represents either surface or canonical pronunciations, and the output is a real vector. Denote by $\vv^c$ and $\vv^s$ the output of the RNNs for the canonical and the surface representations, respectively. Then, both vectors are concatenated $[\vv^c, \vv^s]$ and are fed to three fully connected layers with shared parameters over time. The output is a series of vectors, one for each time step. These are all concatenated\footnote{We apply sequence padding when needed.} and are fed to a fully-connected layer followed by a softmax layer. The similarity function $f$ is the output of the softmax layer, and it can be interpreted as a probability function. The whole network is trained so as to minimize the negative log likelihood loss function. The network architecture is depicted schematically in Figure~\ref{fig:net_arch_1}.

\subsection{Ranking loss function}

\begin{figure}[t]
  \centering
  \includegraphics[width=73mm, scale=0.8]{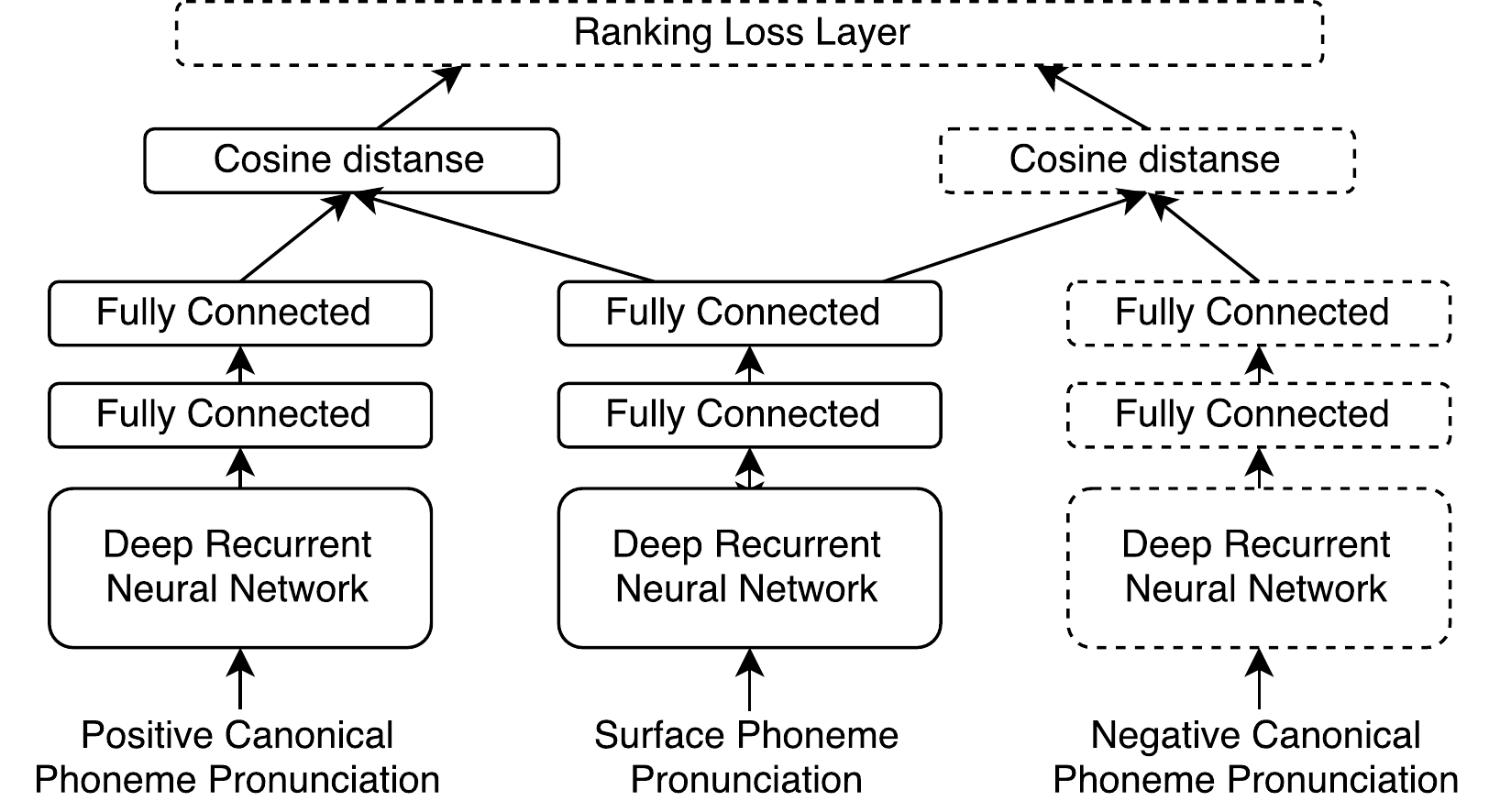}
  \caption{Network architecture of the ranking model. The three RNNs have share parameters and output an embedding vector.}
  \vspace{-.5cm}
  \label{fig:net_arch_w}
\end{figure}

A different approach to learning the similarity function is to score the similarity between a given pronunciation and other pronunciations according to their ranking. In order to do so we use a slightly different training set than the one used to train the binary network. Each example in the new training set is composed of a triplet: a surface pronunciation $\vp^s\in\P$, a positive canonical pronunciation $\vp^+\in\P$, and a negative canonical pronunciation  $\vp^-\in\P$. The positive canonical pronunciation is  the canonical pronunciation associated with the surface pronunciation $\vp^s$, and the negative canonical pronunciation is a canonical pronunciation of a different word. Overall the training set of $m$ examples is denoted $S=\{(\vp^s_i, \vp_i^+, \vp_i^-)\}_{i=1}^{m}$. As in the previous model, we represent each of the three pronunciations using an RNN. The output of the RNN is fed into two fully-connected layers and is considered to be the \emph{pronunciation embedding}. Pronunciation embedding is a function $g$ that maps a pronunciation $\vp\in\P$ to a fixed size vector, $\vu\in\R^n$, where $\vu=g(\vp)$. We measure the closeness between the two embeddings, $\vu\in\R^n$ and $\vv\in\R^n$, using the cosine distance score \cite{KamperWangLivescu16}:
$$
d_{\cos}(\vu,\vv)=\frac{1}{2}\left( 1 -\frac{\vu\cdot\vv}{\|\vu\| \|\vv\|}\right).
$$
The cosine distance score between two embeddings is close to 0 if the vectors $\vu$ and $\vv$ are close, and close to 1 if they are far apart. Formally, the similarity function between two pronunciations $\vp_1$ and $\vp_2$ is defined as 1 minus the cosine distance 
\begin{equation}\label{eq:cos_dis_func}
f(\vp_1, \vp_2) = 1 - d_{\cos}(g(\vp_1), g(\vp_2)).
\end{equation}
During training we minimize the hinge loss over the cosine distance so that the score of the related canonical-surface pronunciations is higher than the score of the unrelated canonical-surface pronunciation by a margin of at least $\gamma$, where $\gamma \in \R_+$ is a positive scalar. Formally, the loss function is defined as \cite{KeshetGrBe09, KamperWangLivescu16}:
\begin{equation}
	\label{eq:rank_loss}
	\ell(\vp^s,\vp^+,\vp^-) = \max\{0, \gamma - f(\vp^s, \vp^+) + f(\vp^s, \vp^-)\}	,
\end{equation}
where $f$ is the similarity function defined in \eqref{eq:cos_dis_func}, and $\gamma$ is the margin parameter. The network architecture is depicted schematically in Figure~\ref{fig:net_arch_w}.

This architecture has three advantages over the binary architecture: (i) the ranking optimization criterion is closer to the notion of similarity function: related pronunciations get higher score than unrelated ones; (ii) as a by-product we introduce the \emph{pronunciation embedding}, which maps a sequence of phones to a fixed size vector; and (iii) we use many negative surface pronunciations for each positive surface pronunciation and increase our training data. We found that it has a great impact on the model's performance, and we analyze it in Section~\ref{sec:ana}.

%% file: 04_exp.tex

\section{Experiments}\label{sec:exp}

We evaluated the proposed architectures on the lexical access task, where we would like to predict the word in the dictionary that is associated with a given surface pronunciation. All experiments are conducted on a subset of Switchboard conversational speech corpus that has been labeled at a fine phonetic level \cite{greenberg:stp}; these phonetic transcriptions are the input to our similarity models. The data subsets, phone set $\mathcal{P}$, and dictionary $\mathcal{V}$ are the same as those previously used in \cite{livescu2004feature, Jyothi2011, hao2012discriminative}. The dictionary contains 5,117 words, consisting of the most frequent words in Switchboard. The base-form uses a similar, slightly smaller phone set (lacking, e.g., nasalization). We used the same partition of the corpus as in \cite{livescu2004feature, Jyothi2011, hao2012discriminative} into 2,942 words in the training set, 165 words in the development set, and 236 words in the test set. 

Results are presented in terms of the word error rate when the top k predictions are considered. This is denoted by WER@k. Table~\ref{tab:1} summarizes the results for all our architectures. For each architecture we tested three types of RNNs:  LSTM with one layer, LSTM with two layers (2-LSTM), and bidirectional LSTM with two layers (BI-2-LSTM). We optimize all our models using Adagrad \cite{duchi2011adaptive} with learning rate value of 0.01. We use ReLU \cite{nair2010rectified} as an activation function after each fully connected layer. For the ranking models we use a margin value of $\gamma=0.3$. All hyper-parameters were tuned on a  validation set. 

\begin{table}[h!]
  \caption{WER for the binary-loss model and the ranking-loss model on the lexical access problem.}
  \label{tab:1}
  \centering
  \begin{tabular}{l l l l}
    \toprule    
    & \multicolumn{1}{c}{\textbf{Models}} & \multicolumn{2}{c}{\textbf{Test}} \\
    & \multicolumn{1}{c}{} & \multicolumn{1}{c}{\textbf{WER@1}} & \multicolumn{1}{c}{\textbf{WER@2}}\\ 
    \midrule
	& dictionary lookup \cite{livescu_thesis} & $59.3\%$  & - \\
    & dictionary + Levenshtein dist. & $41.8\%$  & - \\
    & Jyothi et al., 2011 \cite{Jyothi2011} & $29.1\%$  & - \\
    & Hao et al., 2012 \cite{hao2012discriminative} & $15.2\%$  & - \\  
	\hline        
    \multirow{3}{*}{\rotatebox[origin=c]{90}{Binary}} & LSTM & $20.8\%$  & $18.2\%$ \\
    & 2-LSTM & $24.6\%$  & $22.5\%$ \\
    & BI-LSTM & $21.6\%$  & $19.5\%$ \\   
    \hline
    \multirow{3}{*}{\rotatebox[origin=c]{90}{Rank}} & LSTM & $25.9\%$  & $16.5\%$ \\
    & 2-LSTM & $22.9\%$  & $14.8\%$ \\
    & BI-LSTM & $23.3\%$  & $15.3\%$ \\   
    \bottomrule
  \end{tabular}  
\end{table}

For comparison we added to Table~\ref{tab:1} the word error rate of other algorithms for lexical access: a dictionary lookup with and without Levenshtein distance  \cite{livescu2004feature}, a dynamic Bayesian network (Jyothi \emph{et al.}, 2011 \cite{Jyothi2011}), and discriminative structured prediction model (Hao \emph{et al.}, 2012 \cite{hao2012discriminative}). It can be seen from the table that both of our models outperform the dictionary lookup approaches and the model which is based on dynamic Bayesian networks \cite{Jyothi2011}. However the discriminative structured prediction model \cite{hao2012discriminative} performs much better than any of our models. The discriminative model was trained specifically for the lexical access task with a unique set of feature maps, but it cannot be used as a similarity score between two pronunciations. 

The performance of the binary loss model and the ranking model are almost the same, with the binary loss performing slightly better than the rank loss. The reason is likely that it was trained to maximize the probability that two pronunciations are related and not to match a similarity score. 



%% file: 05_ana.tex

\section{Analysis}\label{sec:ana}

In this section we analyze the performance of the ranking model. We investigate the effect of the number of negative examples, the effect of the embedding size and present some of the model's outputs and the way it errs. 

\vspace{-0.2cm}
\subsection{The effect of the number of negative samples}

Recall that the ranking loss model was trained on triplet made of a canonical pronunciation of a word, a surface pronunciation of the word (positive sample), and a surface pronunciation which is not associated with the word (negative sample). In our experiments the negative surface pronunciation was a surface pronunciation of a random word.

Deep neural networks require a lot of training data in order to converge to a good local  minimum. We expanded the training set by using many different negative samples for each positive samples. In order to examine if this approach leads to a better performance, we trained the network several times with a different number of negative samples per positive sample and evaluated the performance on the test set.

\begin{figure}[t]
  \centering
  \includegraphics[width=68mm, scale=0.9]{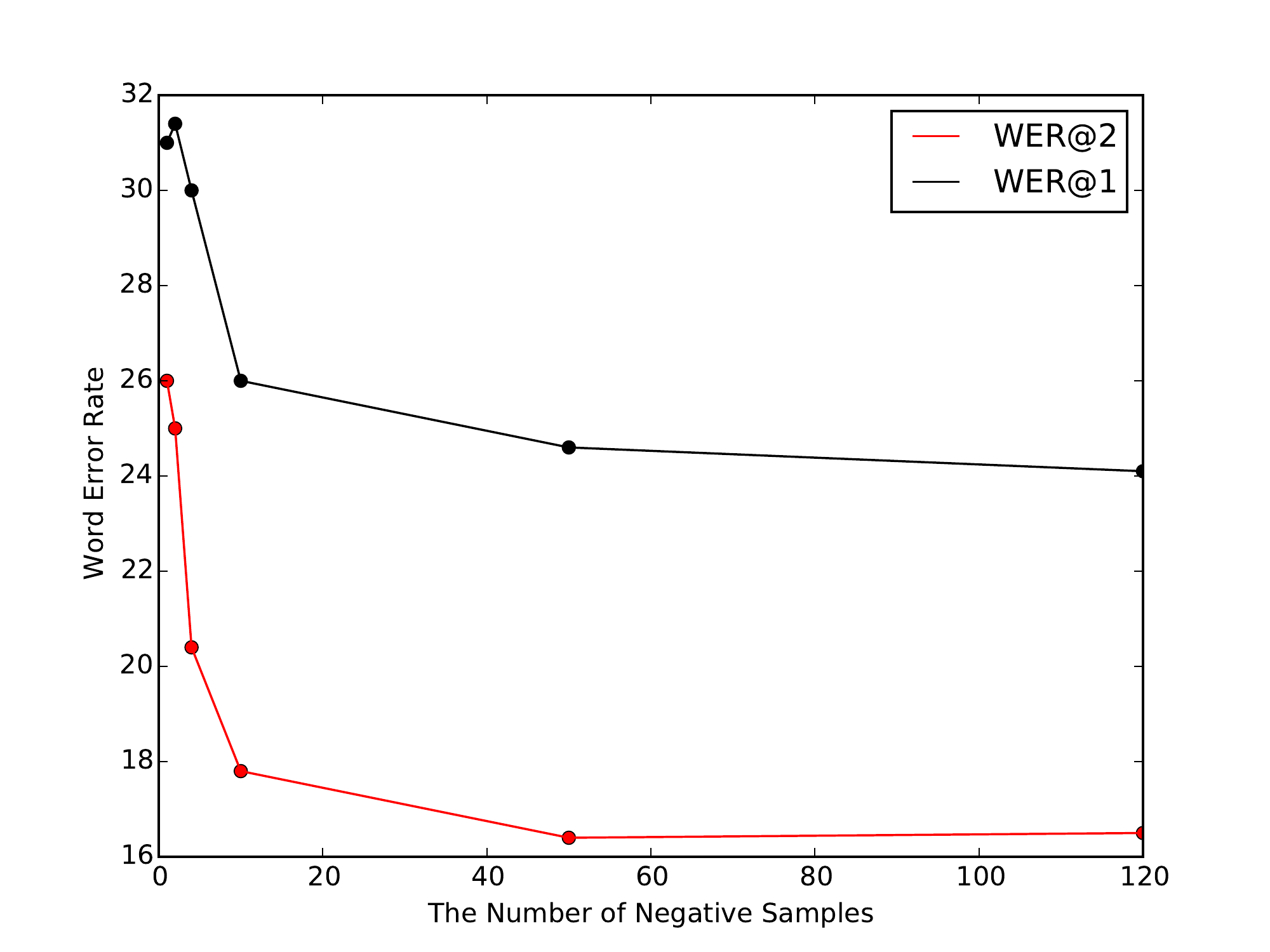}
\vspace{-0.2cm}
  \caption{WER@1 and WER@2 of the test set as a function of the number of samples per example.}
\vspace{-0.2cm}
  \label{fig:n_examp}
\end{figure}

Figure~\ref{fig:n_examp} shows WER@1 and WER@2 of the lexical access task, as a function of the number of negative samples per positive one. Notice that when adding more examples the error rate keeps decreasing until the limit of ~50 negative samples, from this point the error rate stays roughly the same.

\begin{figure}[t]
  \centering
  \includegraphics[width=56mm]{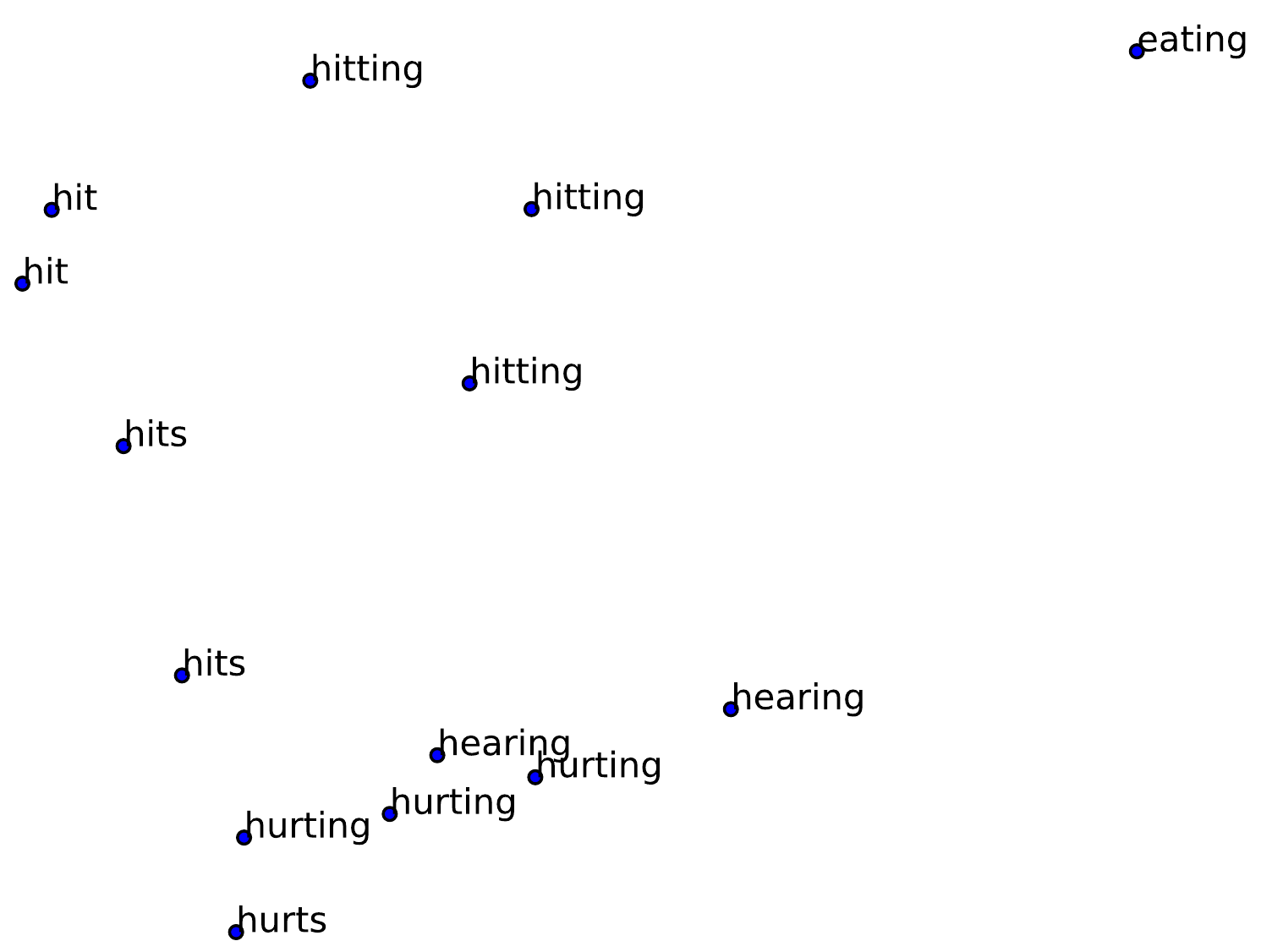}
  \vspace{-0.2cm}
  \caption{Two-dimensional t-SNE projection of the representation vectors (zoom to a specific area.)}
 \label{fig:word_neighborhood}
  \vspace{-0.5cm}
\end{figure}

\subsection{The effect of the embedding size}

Next, we investigated the effect of the pronunciation embedding size $n$, i.e., the size of the output of the last layer after the RNN. We tried different embeddings sizes, and evaluated their performance on the validation and test sets. Table ~\ref{tab:2} summarizes the results. From the table we see that embedding of size $40$ is too small, but the performance of embeddings of size 80 and above are all good. We found the embedding of size  120 to yield the best performance.

\begin{table}[h]
  \vspace{-0.2cm}
  \caption{Performance (WER) for different embedding sizes.}
  \vspace{-0.1cm}
  \label{tab:2}
  \centering
  \begin{tabular}{ l l l l l}
    \toprule
    \multicolumn{1}{c}{\textbf{Dim.}} & \multicolumn{2}{c}{\textbf{Test}} & \multicolumn{2}{c}{\textbf{Validation}} \\
    \multicolumn{1}{c}{} & \multicolumn{1}{c}{\textbf{WER@1}} & \multicolumn{1}{c}{\textbf{WER@2}} & \multicolumn{1}{c}{\textbf{WER@1}} & 
    \multicolumn{1}{c}{\textbf{WER@2}} \\ 
    \midrule
    $150$ & $24.2\%$  & $16.9\%$ & $18.8\%$  & $12.7\%$ \\
    $120$ & $25.9\%$  & $16.5\%$ & $18.2\%$  & $11.5\%$ \\
    $80$ & $25.0\%$  & $17.4\%$ & $21.2\%$  & $14.6\%$ \\
    $40$ & $27.1\%$  & $18.7\%$ & $24.9\%$  & $15.8\%$ \\   
    \bottomrule
  \end{tabular}
  \vspace{-0.2cm}
\end{table}

\subsection{Visualization}

Lastly, we performed a few visualizations in order to get a sense of what the model  learned. In Figure~\ref{fig:word_neighborhood} we visualized a subset from the embedding space of the canonical pronunciations in the dictionary, using t-SNE \cite{maaten2008visualizing} for dimensionality reduction. The words which have a similar pronunciation appear to be close in the embeddings space.

In Table~\ref{tab:word_neighborhood} we present the 4 most similar words according to the learned similarity function for the words \emph{sense, write, die, male}, and \emph{their}. Again, we can see that the most similar words in the embedding space are the words which contain a similar phone sequence.

\begin{table}[h]
\vspace{-0.2cm}
  \caption{Four most similar words from the dictionary computed in the embedding   space.}
\vspace{-0.1cm}
  \label{tab:word_neighborhood}
  \centering
  \begin{tabular}{ l l}
    \toprule
    \textbf{word} & \textbf{neighborhood} \\ 
    \midrule
    sense & cents, since, sent, sentence \\
    write & right, ride, righty, writing \\ 
    die & diet, died, dying, idea \\
    male & mail, meal, may, makes  \\
    their & there, they're, there'd, there're  \\
    
    \bottomrule
  \end{tabular}
\end{table}

In Table~\ref{tab:la_results} we illustrate the predictions of the model for the lexical access task for hard cases of surface pronunciations. The table shows the first three predictions of the model, ordered (left to right) from most similar to least similar. The table is separated into two panels: the upper panel shows the correct predictions made by the model and the lower panel shows incorrect predictions. It can be seen that both the correct and the incorrect prediction are hard to classify, even for a human. In the lower panel, it seems that there might be an error in the transcription (e.g., $[s, tcl, t, ao, r]$ is more similar to the predicted word \emph{store} rather than to labeled word \emph{start}), or that we have reached the limit of the possible discrimination in this task (e.g., $[w_n, ah_n, n]$ can equally be predicted as either \emph{want} or \emph{won}. 

\begin{table}[h]
  \caption{Prediction of the model for hard cases of surface pronunciations.}
  \vspace{-0.2cm}
  \label{tab:la_results}
  \centering
  \footnotesize
  \begin{tabular}{ l l l}
    \toprule
    \textbf{surface pronunciation}& \textbf{desired word} & \textbf{predicted words} \\ 
    \midrule
    $[bcl, b, ao]$ & bought & bought, bob, ball \\
    $[m, ey1, ey2, dcl, jh, er]$ & major & major, mayor, made \\
    $[f, ay1, ay2, n, ih_n, ng]$ & finding & finding, fighting, flying \\ 
    $[pcl, p, r, aa, er]$ & proper & proper, prior, property \\
    $[n, pcl, p, eh, er, z]$ & peppers & peppers, persons, present\hspace{-0.2cm} \\
    \midrule
    $[s, tcl, t, ao, r]$ & start & store, star, sent \\
    $[eh, r, uw, ay1, ay2]$ & everybody & iraq, dry, era \\ 
    $[w_n, ah_n, n]$ & want & won, one, want \\
    $[pcl, p, uw, el]$ & people & pool, pull, people  \\    
    \bottomrule
  \end{tabular}
\end{table}

%% file: 06_conc.tex

\vspace{-.5cm}
\section{Conclusion}\label{sec:conc}

We presented two very different architectures to learn similarity between two phone streams. The first architecture learns the alignment between phone streams that represent close pronunciations. The second architecture is designed to learn a mapping of a phone stream to a vector space, such that close pronunciations will have close representation in the output vector space.

Future work will explore similarity between other sub-word units. Specifically we would like to propose a similarity function between articulatory features, and analyze it in the light of articulatory phonology \cite{browman1992articulatory}. 

\section{Acknowledgment}

 We would like to thank Karen Livescu, Preethi Jyothi and the anonymous reviewers for their helpful comments.